# Weakly Supervised Cross-platform Teenager Detection with Adversarial BERT

Peiling Yi
p.yi@qmul.ac.uk
Queen Mary University of London
London, UK

Arkaitz Zubiaga
a.zubiaga@qmul.ac.uk
Queen Mary University of London
London, UK

## ABSTRACT

Teenager detection is an important case of the age detection task in social media, which aims to detect teenage users to protect them from negative influences. The teenager detection task suffers from the scarcity of labelled data, which exacerbates the ability to perform well across social media platforms. To further research in teenager detection in settings where no labelled data is available for a platform, we propose a novel cross-platform framework based on Adversarial BERT. Our framework can operate with a limited amount of labelled instances from the source platform and with no labelled data from the target platform, transferring knowledge from the source to the target social media. We experiment on four publicly available datasets, obtaining results demonstrating that our framework can significantly improve over competitive baseline models on the cross-platform teenager detection task.

## CCS CONCEPTS

• **Artificial intelligence** → *Natural language processing*; • **Human-centered computing** → Social media.

## KEYWORDS

Natural language processing, Age detection, Deep transfer learning, BERT

## 1 INTRODUCTION

With the proliferation of digital devices and the Internet, social media has become an addictive platform for teenagers [1]. However, long times of online presence can lead to psychological and physical issues, such as worry, depression and social phobia, and even extreme behaviours like suicide [16]. Likewise, teenagers are known to be the main target of online cyberbullying attacks [10]. This leads to the pressing issue of protecting teenagers online [12, 15, 26], for which a crucial first step is to have an ability to detect teenage users in social media.

As a problem linked to author profiling [11], teenager detection can be tackled as a specific case of age detection [13], i.e. binary classification determining if a user is a teenager or not. A common approach is based on the analysis of the author's writing style or content-based characteristics to reveal the author's various attributes, including age [28].

However, teenager detection suffers from scarcity of publicly available datasets and the difficulty of collecting labelled data. Even in research using manual annotations, labelling datasets is costly and prone to biases, and accurate detection is largely bounded by the size and quality of datasets [13]. In addition, with the rapid growth of social media platforms, users move to different platforms very quickly [25]; some of these platforms are more private and labels are more difficult to be retrieved or inferred, and features of social media platforms are substantially different from each other. When it comes to developing a classifier, a model trained on one social media platform will face additional challenges when applied to another platform [29, 32].

To address these challenges, we propose a cross-platform teenager detection framework. Our framework uses a small amount of labelled data from the source social media platform to identify teenagers on the target social media for which labels are not available.

The proposed framework consists of four components: Concentrator, Discriminator, BERT Encoder Measurer and Small datasets adaptive classifier. These four components are designed to improve the small dataset and cross-platform adaptability of BERT [5]. In order to validate the effectiveness of this framework, we evaluate on source-target platform pairs involving four real-world datasets from multiple platforms. We experiment with a small randomly selected subset (700) of training data, in both in-platform and cross-platform settings. The results demonstrate that our framework can significantly improve the performance of the cross-platform teenager detection task using a small amount of source platform instances.

To the best of our knowledge, this is the first work to consider age detection with a very small source dataset to do knowledge transfer on different social media platforms based on a pre-trained model and a deep transfer learning algorithm. We perform a series of ablation studies to objectively evaluate the contribution of each proposed component. This in turn provides new insights into transferring knowledge across social media platforms, extensible to other tasks.

## 2 RELATED WORK
### 2.1 Transfer learning on age detection
Transfer learning can be defined as a method of machine learning that aims to use the knowledge of the source to help improving the prediction function in the target domain [33]. The study of transfer learning for age detection is not a new research topic and was pioneered by Nguyen et al. [14]. This study applied the feature augmentation method [4] on a linear regression model to address cross-domain age detection on three distinct data genres simultaneously: blogs, telephone conversations and online forum posts.

PAN is a series of shared tasks on author profiling which has been running since 2013 [18]. In 2014 [17], PAN launched a shared task including age detection, with a training dataset containing four different genres (social media, blog, Twitter and forum conversations) to analyse the adaptability of these detection approaches. A general model is trained to test on four sub-datasets, which worked best on the Twitter English dataset (63%) for the age detection sub-task and performed poorly on other English datasets (below 40%) [17]. In 2016, PAN used the Twitter corpus as a training dataset to set up another interesting job. The challenge is that the model is evaluated on a testing dataset completely different from the training dataset. The best result (58.97%) in this competition using stylistic and second-order features into SVM [20]. Still, none of the participants used a transfer learning algorithm to build their models. The PAN shared tasks did not address transfer learning for the age detection task in subsequent years.

### 2.2 Adversarial domain adaptation network
Adversarial domain adaptation network [8] is motivated by Generative Adversarial Networks (GAN) [9]. GAN consists of two parts: (1) the Generator, responsible for generating synthetic instances; and (2) the Discriminator, responsible for judging whether a sample is real or artificially generated. The game between the generator and the discriminator completes the confrontation training. The purpose is to map the source domain input and target domain input into the same feature space. Then, the classifier trained on the source domain (with labels) can be directly used for the classification of the target domain data. Inspired by the GAN mechanism, Tzeng et al. [27] proposed Adversarial Discriminative Domain Adaptation (ADDA) [27], which solely adopts the discriminator component to make the training more efficient; Our proposed framework leverages ADDA's discriminator.

### 2.3 BERT fine-tuning using small datasets
Bidirectional Encoder Representations from Transformers (BERT) [5] is a transformer-based machine learning technique, which can help many downstream tasks of natural language processing achieve breakthrough performance. Compared with deep training models trained from scratch, the main advantage of BERT is that they can be adapted to specific tasks by using a relatively small amount of labelled data to get a robust model. Recent work has explored how to adjust the network architecture and hyper-parameter to fit small datasets [7, 23, 31].

Our work is inspired by these findings but focuses on how to improve cross-platform performance when the source dataset is very small. The inherent limitation of BERT's fine-tuning is that it cannot handle the situation where the distribution of training data and test data are different, which we address here.

## 3 METHOD
### 3.1 Problem definition
In this study, we define teenager detection as a binary classification task consisting in determining if each text in $T$ $T \in \{., T_n\}$ pertains to a teenager, i.e. $y \in \{0, 1\}$, where $y = 1$ indicates a teenager (younger than 20) and $y = 0$ indicates an adult (20 or older). Datasets belong to the source platform $s$ or target platform $t$, which leads to two different input spaces $X_s$ and $X_t$ where $X_s \neq X_t$, but the same label space $Y_s == Y_t$. Moreover, the data distribution of the source platform is $P_s(x, y)$, which is different from the data distribution of the target platform $P_t(x, y)$ and these two distributions are both unknown and imbalanced. We aim to leverage a function $M$ which can map input spaces $X_s$ and $X_t$ into a single, common space $X$.

How much knowledge of the source platform can be transferred to the target platform depends on three factors: 1) hypothesis loss in the source domain; 2) the divergence between the target platform presentation and source platform presentation; and 3) the loss from the classifier model across the platforms. When the combined loss is large, the model would struggle to perform well across platforms. Our framework incorporates four components that aim to minimise these losses.

Given the scarcity of labelled datasets, we tackle it as a weakly supervised task where a model $C_s$ is trained from little labelled source data to detect teenagers on the target platform lacking labelled data for training.

### 3.2 Model Architecture
*3.2.1 Concentrator.* BERT can keep up to 512 tokens as input, and it is generally recommended to truncate the input content [24]. However, for knowledge transfer on platforms with very different text lengths, this method tends to miss potentially important information in the input. Thus we propose the Concentrator component. The function of the Concentrator is to extract important fragments from all target and source platform inputs using feature engineering. The Concentrator attempts to align all inputs before training. How many tokens will be kept in the source platform depends on the length of the input from the target platform. In addition, we combine an external age prediction lexicon [21] with the unigram and bigram usage in all platforms to get a 900 items lexicon. The lexicon is used to align the input tokens in different platforms.

*3.2.2 Source BERT & Target BERT.* In this research, the BERT model is used as the sentence embedding encoder. All input sentences going through the BERT model will be mapped to an input space $X$, where each $x \in X$ is a 768-dimensional vector. Source BERT and target BERT were trained at different points in time. First of all, the source BERT model is generated by standard supervisor training. Secondly, the parameters of trained source BERT are used to initialise the target BERT.

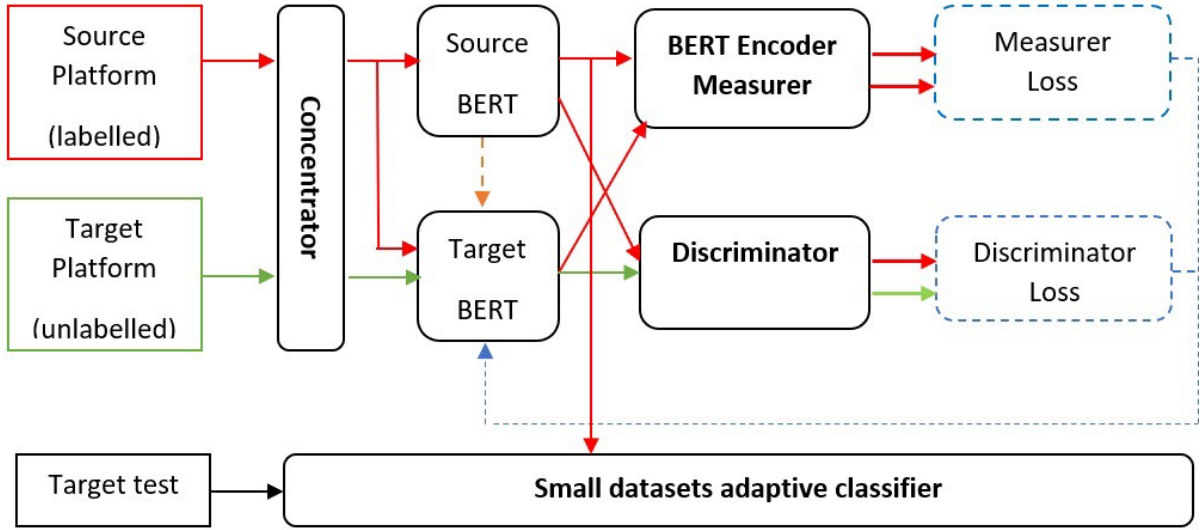

**Figure 1: Model Architecture.** The red line represents the source platform data flow. The green line represents the target platform data flow. The blue dashed line shows how the loss is fed back to the backpropagation algorithm. The orange dashed line is the source BERT parameters, which are used to initialise the target BERT.

*3.2.3 Discriminator & BERT Encoder Measurer.* The discriminator component is based on the ADDA framework [27]. In this study, the learning goal is: the trained target BERT can map the target input representation to the source input space, making it difficult for the discriminator to determine the platform that the input comes from. Discriminator consists of two fully-connected layers on top of BERT encoders, finished with a sigmoid activation. A supervised loss function for discriminator component is defined as follows:

$$min_{adv_D}(X_s; X_t; M_s; M_t) = -E_{x_s} \sim X_s[logD(M_s(x_s))] \\ - E_{x_t} \sim X_t[log(1 - D(M_t(x_t)))] \quad (1)$$

During the experiments, we observed that when the training dataset is very small, gradient vanishing is common. To solve this, another component is added: BERT Encoder Measurer. It measures the differences between source BERT and target BERT. The learning goal is to get a similar hypothesis when the target BERT encoder and the source BERT encoder face the same source dataset. The loss function adopts the Kullback–Leibler divergence [30] to minimize the difference of two probability distribution can be defined as follows:

$$minL_{ts}(X_s; M_s; M_t) = E_{x_s} \sim X_s[M_s(x_s)](log(E_{x_s} \sim X_s[M_s(x_s)]) \\ - E_{x_s} \sim X_s[M_t(x_s)]) \quad (2)$$

Then, we use two losses (discriminator loss and BERT encoder measurement loss) to train the target BERT model.

*4.1.1 Small datasets adaptive classifier.* This component is a classifier trained on source data that can be applied to the target data. To make the input of the last layer of the classifier denser so that it can accommodate a smaller dataset, we use a two-layer feedforward network with ReLU activation and 512 hidden sizes for the first layer and Softmax activation for the output layer.

## 4 EXPERIMENTS

For evaluation, we design our experiments to answer the following research questions:

- **RQ1:** How effective is the proposed framework for teenager detection across platforms?

- **RQ2:** How does each component affect the entire framework?

- **RQ3:** How to choose the source platform dataset to help knowledge transfer?

### 4.2 Datasets

We use four publicly available datasets for our experiments (see Table 1). Two of the datasets were designed for age detection (Blogger and Pan13), whereas the other two were designed for cyberbullying detection (YouTube and Myspace). In both cases, we adapt the age labels provided in the datasets for the teenager detection task. Pan13 differs from the rest of the datasets in that it contains data from multiple platforms; given our focus on cross-platform teenager detection, Pan13 is only used as a target dataset in our experiments, as otherwise we would not be able to control the platforms considered for training.

It is worth noting that the average text length of each platform varies greatly. Fixing the input length to the average text length is a possible solution to keep the maximum distinctive information when training BERT model. However, in cross-platform training, we manage to make a trade-off between maximising distinctive information and reducing vector sparsity. In addition, all the datasets except blogger are imbalanced. This reality makes even in-platform model training a big challenge.

### 4.3 Setup

We test our model in weakly supervised settings using smaller train- ing samples, which was set to stratified samples of 700 instancesafter empirical testing.

In addition to experiments testing our proposed framework, we also experiment with a baseline model that leverage full training sets using a BERT base model [5]. In setting up the BERT model, we rely on the training hyper-parameters recommended by [24]; Batch size: 16; Learning rate (Adam): 2e-5; Number of epochs: 4.

We report Macro F1 performance scores for all experiments.

### 4.4 Results

Table 2 shows the Macro F1-score of different models generated by our proposed framework on the cross-platform settings. Numbers in bold indicate values exceeding the baseline. Bold and underlined indicate the best result(s) for each source-target dataset setting. In what follows we analyse the results in relation to the three research questions set forth in this work.

*Effectiveness of the proposed framework (RQ1).* We observe that models based on our proposed framework outperform the cross-platform baseline model, except for one case (M->Y), demonstrating the effectiveness of our weakly supervised framework using only 700 instances for training, as opposed to the full dataset used by the baseline model. The model incorporating all components, *AB_CSA*, is the best model overall. To intuitively evaluate the effect of non-adversarial network components applied to BERT, we further performed in-platform testing (see Table 3). Despite that our model only uses 700 training instances, it can still outperform the in-platform baseline except for the MySpace dataset.

*Ablation studies (RQ2).* Aiming to gain a better understanding of the contribution of each component on the overall system, we test by removing each of the components.

We find that when only *AB_CS* and *AB_C* are used to fine-tune on the source and without utilising any target data, the model can still outperform the baseline. This validates the effectiveness of our proposed pre-training component: Concentrator. During the training process, we aim to reduce the distance between the input space of the source platform and the target platform. Therefore, a denser classification layer is established. The average performance of *AB_S* is 2% higher than the baseline. To our surprise, only using the Adversarial Adaption network (*AB_A*) fails to meet our expectations. Even on some platforms, it can degrade the overall performance.

Just like the theory from [3], we need high-quality source datasets to store knowledge, appropriate methods to reduce the difference between source data and target data, and classifier with less loss to help transmission.

Assessing the impact of each component on the model, we find the following:

- **Impact on Concentrator.** An attractive feature of our framework is the use of Concentrator, which can not only enrich the limited token information but also shorten the distance between the two platform datasets easily and cheaply. The final experimental results show that the greater the difference in text length distribution, the more obvious the role of Concentrator. In some cases, only using a Concentrator can achieve the best results.

- **Impact on Adversarial Adaption network.** This is the first study to combine BERT with Adversarial Adaption network and a small number of samples for age detection task. According to our observations, Adversarial Adaption network alone is difficult to learn how to map the target input representation to the source input space on very different platforms.However, in these datasets with little difference in data distribution, Adversarial Adaption network can make the greatest contribution.

- **Impact on Small datasets adaptive classifier.** For text classification, it is recommended to add a fully connected output layer to finetune the pre-trained BERT model [24]. However, it may fail to distinguish when applied to scenarios with big datasets shift in small datasets. The small datasets adaptive classifier aims to reduce the distribution difference between the source and the target data via reducing the dimensionality of input representations. We note that this component is particularly useful when the average input length between the target and the source platforms varies greatly.

*Source platform selection (RQ3).* Combining the results in Ta- ble 2 and Table 3, we can see that data quality is more importantthan data size when it comes to training data. The knowledge trans-fer is substantially impacted by the quality of the source platform as well as the similarities between source and target platforms. For example, the Blogger dataset has good in-platform performance, soas a source platform, it can provide the best cross-platform results (B->Y, B->M and B->P).

On the contrary, due to the poor in-platform performance of the Myspace dataset, it is not not the best option to store the knowledge required for transfer learning, which will greatly increase the error on the target predictions. In addition, compared with other platforms, each input text of the Myspace dataset contains the least information (the average input length is 17). Therefore, it is difficult to map all inputs to the common input space. This means that the reduction in errors is always small because the source data is too far away from the target data. The interesting phenomenon is that when MySpace is used as the target platform (B->M,Y->M), performance can be greatly improved.

## 5 CONCLUSION

To tackle the teenager detection task in the absence of labelled data for some social media platforms, we propose a novel weakly supervised cross-platform framework, which improves the small data and cross-platform adaptive capabilities of BERT. Our framework leverages a dual input alignment strategy, which takes into account the input data space and the latent representation space, to reduce the large dataset shift between different platforms. In addition, we design a Small datasets adaptive classifier and a BERT Encoder Measurer to stimulate BERT and Adversarial Adaption network's ability to adapt to small datasets.

| Platforms | YouTube | Myspace | Blogger | PAN13 (Netlog, Blogspot, Internetwordstats) |
|---|---|---|---|---|
| **Size** | 3,468 | 14,813 | 19,320 | 236,600 |
| **Avg. length** | 115 | 17 | 3766 | 505 |
| **TR** | 0.2 | 0.096 | 0.42 | 0.08 |
| **Year** | 2020 | 2011 | 2009 | 2013 |
| **Source** | Elsafoury [6] | Bayzick and Kontostathis [2] | Schler et al. [22] | Rangel et al. [19] |

Table 1: Dataset statistics. TR: teenager ratio, as the portion of users in the dataset that are labelled as teenagers.

| | Baseline | Full model | Ablated models | | | | | |
|---|---|---|---|---|---|---|---|---|
| Source->Target | BERT | AB_CSA | AB_CS | AB_S | AB_C | AB_A | AB_CA | AB_SA |
| B->Y | 0.45 | **0.54** | 0.50 | 0.52 | **0.54** | 0.51 | 0.52 | 0.41 |
| B->M | 0.55 | **0.58** | 0.54 | **0.58** | 0.54 | 0.48 | 0.49 | 0.53 |
| B->P | 0.48 | **0.52** | 0.52 | 0.52 | 0.50 | 0.50 | 0.51 | 0.51 |
| Y->B | 0.37 | **0.61** | 0.59 | 0.49 | 0.41 | 0.62 | **0.64** | 0.62 |
| Y->M | 0.49 | **0.53** | **0.54** | **0.54** | 0.48 | 0.49 | 0.49 | **0.53** |
| Y->P | 0.47 | **0.51** | 0.48 | 0.48 | 0.48 | 0.50 | 0.50 | 0.49 |
| M->B | 0.50 | 0.45 | 0.49 | 0.37 | **0.53** | 0.37 | 0.37 | 0.38 |
| M->Y | **0.54** | 0.53 | 0.53 | 0.53 | 0.53 | 0.48 | 0.45 | 0.45 |
| M->P | 0.50 | 0.50 | **0.52** | 0.47 | 0.50 | 0.49 | 0.50 | 0.50 |
| Average | 0.48 | **0.53** | 0.52 | 0.50 | 0.50 | 0.49 | 0.49 | 0.49 |

Table 2: Cross-platform results. B: Blogger dataset; Y: Youtube dataset; M: Myspace dataset; P: Pan13 dataset; AB_* refers to our model incorporating * components, where C is the concentrator, A is the adversarial network component and S is the small dataset adaptive component.

| Source->Target | BASE_LINE | AB_C | AB_S | AB_CS |
|---|---|---|---|---|
| B->B | 0.86 | 0.84 | 0.83 | **0.87** |
| Y->Y | 0.59 | **0.60** | 0.52 | 0.54 |
| M->M | **0.48** | 0.43 | 0.43 | 0.43 |
| P->P | 0.49 | 0.49 | **0.59** | 0.59 |

Table 3: In-platform results

Our experimental results on four real-world datasets show that the framework can significantly improve the performance of in-platform and cross-platform learning on BERT by using small training datasets. To better understanding each component's contribution, we analyse and evaluate the impact of different components on the overall system, which demonstrates the effectiveness of both the input alignment strategy and the small dataset adaptive design. We also find that if the target dataset is of low quality, avoiding use of the target data for training and only using pre-trained components on a different source dataset can achieve better performance.

Our plans for future work include investigation of the applicability of our framework to other tasks and other pre-trained models, and conducting extensive statistical significance testing to further verify the effectiveness of this method. Exploring the impact of noise examples in cross-platform classification is also one of the tasks we will challenge in the future.